\documentclass[10pt,twocolumn,letterpaper]{article}

\usepackage[pagenumbers]{wacv}

\usepackage[utf8]{inputenc}
\usepackage[T1]{fontenc}

\usepackage{url}
\usepackage{booktabs}
\usepackage{amsfonts}
\usepackage{amsmath}
\usepackage{amssymb}
\usepackage{microtype}
\usepackage{xcolor}
\usepackage{graphicx}
\usepackage{subcaption}
\usepackage{flafter}
\usepackage{placeins}
\usepackage{algorithm}
\usepackage{algorithmic}
\usepackage{enumitem}
\usepackage{array}

\definecolor{wacvblue}{rgb}{0.21,0.49,0.74}
\usepackage[breaklinks,colorlinks,allcolors=wacvblue]{hyperref}
\hypersetup{
  pdftitle={Improving Weak World Models Behind Strong Agents in Atari Pong},
  pdfauthor={Yukuan Lu, Zaishuo Xia, Weyl Lu, and Yubei Chen},
  pdfsubject={Visual world models and model-based reinforcement learning}
}

\newcommand{\methodname}{CGSReg}

\newcommand{\mainNativeZeroShotRef}{\autoref{sec:native-zero-shot-mbrl}}
\newcommand{\mainAtariGapFigureRef}{\autoref{fig:atari100k-frozen-wm-summary}}
\newcommand{\mainCGSRegSectionRef}{\autoref{sec:cgsreg}}
\newcommand{\mainUnifiedZeroShotRef}{\autoref{sec:unified-pixel-space-zero-shot-mbrl}}
\newcommand{\mainNativeZeroShotTableRef}{\autoref{tab:native-zero-shot-mbrl-in-pong}}

\def\WACVVERSION{1}

\title{Improving Weak World Models Behind Strong Agents in Atari Pong}
\author{Yukuan Lu \qquad Zaishuo Xia \qquad Weyl Lu \qquad
Yubei Chen\textsuperscript{\textdagger}\\
UC Davis}
\date{}

\begin{document}

\twocolumn[{%
  \renewcommand\twocolumn[1][]{#1}%
  \maketitle
  \vspace*{23.17pt}
  \ifdefined\WACVVERSION
  \begin{minipage}{\linewidth}
    \centering
    \includegraphics[width=\linewidth]{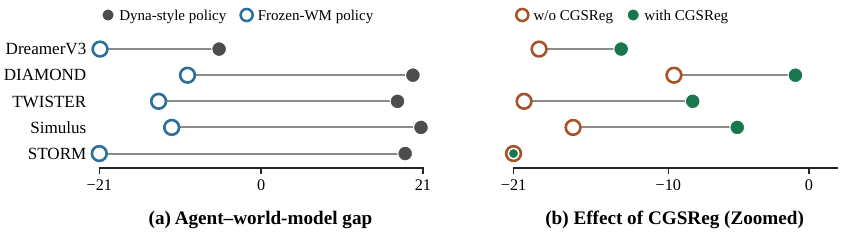}
    \captionof{figure}{Main results in Atari Pong.
    \textbf{(a)} Retraining the policy from scratch inside each reproduced Dyna-style agent's
    frozen world model (WM) causes large drops in game return.
    \textbf{(b)} Under matched offline training, \methodname{} improves
    pixel-space zero-shot MBRL in four of the five evaluated models.
    Higher Pong return is better; panel (b) zooms to $[-21,2]$ from the full $[-21,21]$ range.}
    \label{fig:main-results}
  \end{minipage}
  \vspace{0.5em}
\else
  \noindent\begin{minipage}{\linewidth}
    \centering
    \includegraphics[width=\linewidth]{figures/front_page_results.pdf}
    \captionsetup{type=figure,hypcap=false}
    \caption{Main results in Atari Pong.
    \textbf{(a)} Training the policy from scratch inside each reproduced Dyna-style agent's
    frozen world model (frozen WM) causes large drops in game return.
    \textbf{(b)} Under matched offline training, \methodname{} improves
    pixel-space zero-shot MBRL in four of the five evaluated models.
    Higher Pong return is better; panel (b) zooms to $[-21,2]$ from the full $[-21,21]$ range.}
    \label{fig:main-results}
  \end{minipage}
\fi

  \par\noindent\makebox[\linewidth][l]{%
    \begin{minipage}{\dimexpr\linewidth/2-\columnsep/2\relax}
      \centering\large\bfseries Abstract
    \end{minipage}%
  }
  \vspace{12pt}
}]

\begingroup
  \renewcommand{\thefootnote}{\fnsymbol{footnote}}%
  \footnotetext[2]{Corresponding author.}%
\endgroup

{\itshape\noindent
Strong world-model agents frequently contain weak world models.
We study this agent--world-model gap by reproducing five visual world-model agents in Atari Pong:
DreamerV3, DIAMOND, TWISTER, Simulus, and STORM, with performance comparable
to the reported results, and independently evaluating their frozen world models.
First, closed-loop rollout diagnosis qualitatively inspects visual trajectories
generated by each frozen model under an independently trained policy. All five
models exhibit clear visual or dynamical failures, including ball disappearance,
incorrect motion, and invalid ball--paddle interactions.
Second, under native zero-shot model-based reinforcement learning (MBRL), a new policy is trained entirely
within the frozen model from scratch using the agent's native RL procedure,
without real-environment training. When evaluated in the real environment, these policies
substantially underperform the reproduced agents: DreamerV3
($-5.5 \rightarrow -20.9$), DIAMOND ($19.7 \rightarrow -9.6$), TWISTER
($17.7 \rightarrow -13.3$), Simulus ($20.8 \rightarrow -11.6$), and STORM
($18.7 \rightarrow -21.0$), where $-21$ is the minimum Pong return.
This gap also extends broadly across Atari100K.
Motivated by the ball-related rollout failures in Pong, we propose
Concept-Guided Spatial Regularization (\methodname{}), an auxiliary
reconstruction loss on task-critical concept regions. We evaluate it under a more challenging
pixel-space zero-shot MBRL setting, where policies learn directly from
images generated by the frozen world model. Ball-region \methodname{} improves
pixel-space zero-shot MBRL in DreamerV3, DIAMOND, TWISTER, and Simulus, and
also improves closed-loop rollouts in the first three; STORM shows no clear
improvement.
\par
\vspace{12pt}}

\ifdefined\WACVVERSION
\begin{figure*}[t!]
\else
\begin{figure}[!ht]
\fi
  \centering
  \captionsetup{skip=6pt}
  \ifdefined\WACVVERSION
    \makebox[\textwidth][c]{%
      \includegraphics[width=1.08\textwidth]{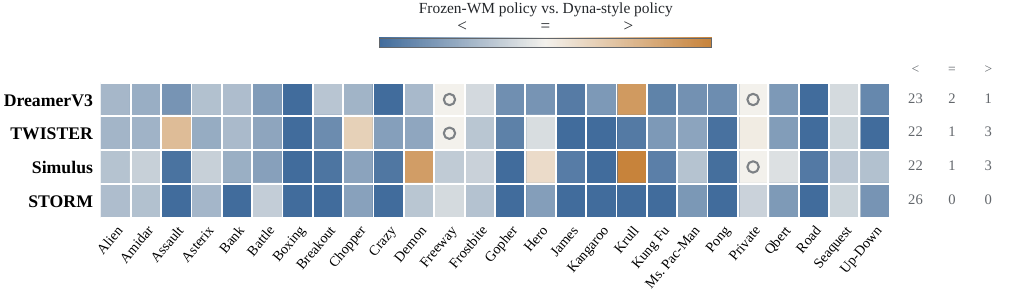}%
    }
  \else
    \includegraphics[width=\linewidth]{figures/atari100k_frozen_wm_summary.pdf}
  \fi
  \caption{\textbf{Agent--world-model gap across Atari100K.}
  Extension of the Pong comparison in \autoref{fig:main-results}(a) to all 26 Atari100K games.
  Across all four projects, Frozen-WM policies underperform the corresponding
  reproduced agents in most games. The mean-return difference is human-normalized and reflected by color intensity, while blue and orange indicate  lower and higher
  performance, respectively. Exact ties are marked with gray circles. DIAMOND is omitted due to computational cost.}
  \label{fig:atari100k-frozen-wm-summary}
\ifdefined\WACVVERSION
\end{figure*}
\else
\end{figure}
\fi

\section{Introduction}
\label{sec:introduction}

World-model agents can achieve strong model-based reinforcement learning (MBRL)
performance without necessarily learning reliable standalone world models.
We refer to this strong-agent/weak-world-model mismatch as the
\emph{agent--world-model gap} and study it in five visual world-model agents---DreamerV3,
DIAMOND, TWISTER, Simulus, and STORM~\citep{hafner2023dreamerv3,alonso2024diamond,burchi2025twister,
cohen2025simulus,zhang2023storm}. We reproduce their Atari Pong training pipelines at comparable performance and freeze the learned world models for independent evaluation.

The frozen models produce visibly incorrect trajectories,
including disappearing balls, unexplained direction changes, and invalid
ball--paddle interactions. More importantly, when a new policy is trained from scratch
inside the final learned world model, with the original RL setup otherwise preserved,
the resulting policy performs far worse in the real environment.
For example, DreamerV3 drops from a mean Pong return of $-5.5$ to $-20.9$,
while STORM drops from $18.7$ to the minimum return of $-21.0$ (\autoref{fig:main-results}(a),
\autoref{tab:native-zero-shot-mbrl-in-pong}).

The ball-related rollout failures further motivate our hypothesis
that task-critical concepts, such as the Pong ball, may receive insufficient learning signal during
world-model training. We propose Concept-Guided Spatial Regularization (\methodname{}), an
auxiliary reconstruction objective that strengthens supervision on segmented
concept regions. \methodname{} improves both closed-loop rollouts and
pixel-space zero-shot MBRL in DreamerV3, DIAMOND, and TWISTER, and improves
pixel-space zero-shot MBRL in Simulus; STORM shows no clear improvement (\autoref{fig:main-results}(b),
\autoref{tab:unified-pixel-space-zero-shot-mbrl-in-pong}).

The gap extends beyond Pong: across Atari100K~\citep{kaiser2020simple}, Frozen-WM policies underperform the reproduced agents in 23/26, 22/26, 22/26, and 26/26 games for DreamerV3, TWISTER, Simulus, and STORM, respectively (\autoref{fig:atari100k-frozen-wm-summary}). This finding echoes MoSim's related observation that RL performance need not reflect world-model predictive capability~\citep{hao2025mosim}.

\ifdefined\WACVVERSION
\else
  \ifdefined\WACVVERSION
\begin{figure*}[t!]
\else
\begin{figure}[!ht]
\fi
  \centering
  \captionsetup{skip=6pt}
  \ifdefined\WACVVERSION
    \makebox[\textwidth][c]{%
      \includegraphics[width=1.08\textwidth]{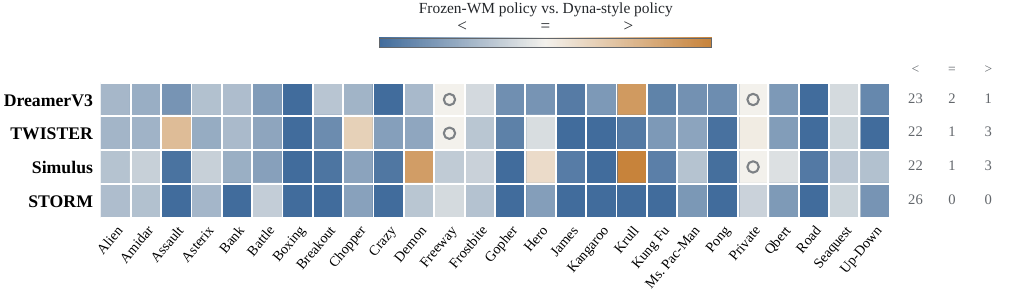}%
    }
  \else
    \includegraphics[width=\linewidth]{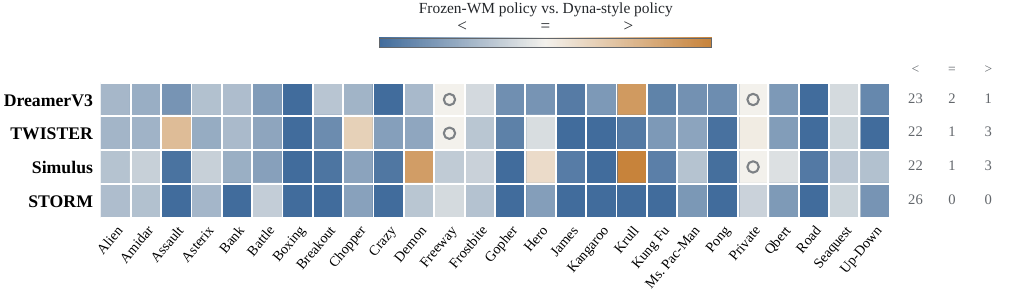}
  \fi
  \caption{\textbf{Agent--world-model gap across Atari100K.}
  Extension of the Pong comparison in \autoref{fig:main-results}(a) to all 26 Atari100K games.
  Across all four projects, Frozen-WM policies underperform the corresponding
  reproduced agents in most games. The mean-return difference is human-normalized and reflected by color intensity, while blue and orange indicate  lower and higher
  performance, respectively. Exact ties are marked with gray circles. DIAMOND is omitted due to computational cost.}
  \label{fig:atari100k-frozen-wm-summary}
\ifdefined\WACVVERSION
\end{figure*}
\else
\end{figure}
\fi

\fi

A world model predicts future observations from past observations and actions.
In MBRL, such learned models are commonly
used as simulated environments to generate trajectories for policy training. The world models of the five agents studied here span latent-state, pixel-space, and discrete-token prediction spaces, with RSSM (DreamerV3), diffusion (DIAMOND), Transformer (TWISTER/STORM), and RetNet (Simulus) backbones. Despite these architectural differences, all follow a broad Dyna-style training pattern that alternates real data collection, world-model learning, and model-based policy improvement~\citep{sutton1990integrated,sutton1991dyna}; we call them \emph{Dyna-style agents}.

We focus on Atari Pong in the Arcade Learning Environment
(ALE)~\citep{bellemare2013ale} under the Atari100K setting.
The agent controls the right paddle against
a computer-controlled left paddle. A player scores when the opponent misses
the ball, and the first side to 21 points wins, giving an episodic return in
$[-21,21]$. \autoref{fig:pong-environment} provides an overview of Pong.

\ifdefined\WACVVERSION
\begin{figure*}[t!]
\else
\begin{figure}[!ht]
\fi
  \centering
  \ifdefined\WACVVERSION
    \includegraphics[width=0.90\textwidth]{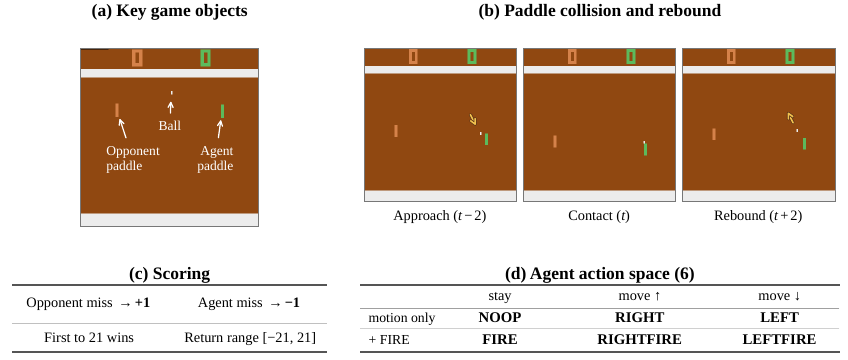}
  \else
    \includegraphics[width=\linewidth]{figures/pong_environment.pdf}
  \fi
  \caption{\textbf{Atari Pong at a glance.}
  (a) Game objects and control roles.
  (b) Ball--paddle interaction; yellow arrows indicate ball motion.
  (c) Scoring after a missed return.
  (d) Six ALE actions: three paddle motions, each with or without
  \textsc{fire}, which starts play after reset.}
  \label{fig:pong-environment}
\ifdefined\WACVVERSION
\end{figure*}
\else
\end{figure}
\fi

We evaluate frozen world models along two complementary dimensions.
The first is \emph{closed-loop rollout diagnosis}, which examines generated
visual trajectories for visual and dynamical failures. The second is
\emph{zero-shot MBRL}, where a policy is trained entirely inside a fixed world
model and then evaluated in the real environment. Here, \emph{zero-shot}
means that the policy receives no learning signal from the real environment, either directly or through further world-model updates.

We use two zero-shot MBRL evaluation protocols:

\begin{itemize}[leftmargin=*, itemsep=2pt, topsep=2pt]

  \item \textbf{Native zero-shot MBRL.}
  We preserve each reproduced agent's native policy-learning procedure to
  diagnose its final frozen world model. Depending on the agent, policy learning may occur in pixel space
(DIAMOND), latent space (DreamerV3, TWISTER, and STORM), or latent-token space
(Simulus).

  \item \textbf{Unified pixel-space zero-shot MBRL.}
  \emph{Pixel-space zero-shot MBRL} is an extension of
  MoSim, replacing physical vector states with pixel
  observations and requiring policies to learn directly from images generated by
  the world model. We further use the same RL setting across architectures,
  yielding a unified protocol for controlled evaluation of \methodname{}.

\end{itemize}

Our contributions are:

\begin{itemize}[leftmargin=*]

  \item In Atari Pong, we isolate and directly evaluate the frozen world models
of five reproduced Dyna-style agents, revealing a clear agent--world-model gap.

  \item We propose \methodname{}, which strengthens reconstruction of
  task-critical concept regions. In Pong, it improves pixel-space zero-shot
  MBRL in four architectures and closed-loop rollouts in three.

  \item We extend the frozen-WM evaluation beyond Pong across the 26-game
  Atari100K benchmark and show that this agent--world-model gap is
  widespread across tasks.

\end{itemize}

\section{Revealing the Agent--World-Model Gap}
\label{sec:revealing-the-agent--world-model-gap}

We reveal the agent--world-model gap by independently evaluating the frozen
world models obtained from the five reproduced Dyna-style agents introduced
above. Each reproduction uses a single training
seed and achieves Pong performance comparable to the corresponding reported
result (\autoref{fig:main-results}(a),
\autoref{tab:native-zero-shot-mbrl-in-pong}). We freeze the learned world model
and evaluate it independently throughout this section.

We use Pong as a controlled setting for these diagnostics. Its dynamics depend
on small objects and precise ball--paddle interactions, making visual and
dynamical errors easy to inspect.

\subsection{Closed-Loop Rollout Diagnosis}
\label{sec:reproduced-wm-closed-loop-rollout-diagnosis}

We generate closed-loop visual rollouts from these five frozen reproduced-agent
world models by pairing each with the same external pixel-based policy. At each step,
the policy selects an action from the model-generated frame, and the
world model generates the next frame. We qualitatively inspect the resulting
trajectories for errors in object visibility, ball motion, collisions, and
action response.

We use an external controller because each agent's original policy is jointly
optimized with its world model. The original policies produce visually
consistent rollouts, whereas the independent controller exposes failures in
the learned dynamics and visual generation. This evaluation therefore probes
the standalone simulation quality of each frozen world model.

The controller is a PPO policy~\citep{schulman2017ppo} trained in real Atari
Pong with Stable-Baselines3~\citep{raffin2021sb3}, using the publicly available
\texttt{PongNoFrameskip-v4} checkpoint.
It is independent of the policies in the reproduced Dyna-style agents.

\autoref{fig:closed-loop-rollout-failures} shows representative failures in
object visibility, ball dynamics, and action response. These failure types
recur across models; for example, TWISTER also exhibits local visual artifacts
around the ball. Their prominence motivates our later focus on the Pong ball
as a task-critical concept.

\ifdefined\WACVVERSION
\begin{figure*}[t!]
\else
\begin{figure}[t]
\fi
  \centering
  \ifdefined\WACVVERSION
    \includegraphics[width=0.90\textwidth]{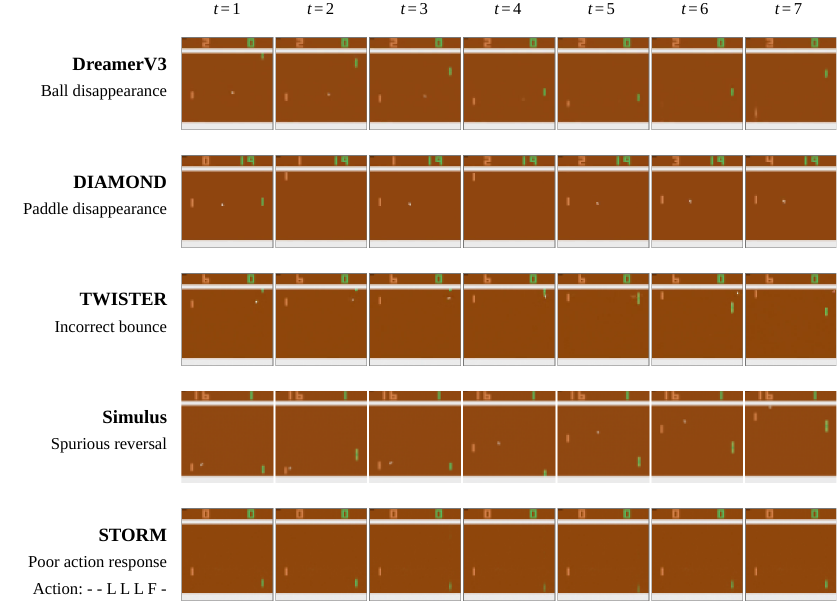}
  \else
    \includegraphics[width=\linewidth]{figures/closed_loop_rollout_failures.pdf}
  \fi
  \caption{Representative closed-loop rollout failures across five frozen
  world models driven by the same external policy. Each row shows a
  seven-frame window, indexed $t=1,\ldots,7$, from a horizon-512 rollout: ball
  disappearance
  (DreamerV3), player-paddle disappearance (DIAMOND), invalid ball--paddle
  interaction (TWISTER), spurious ball-direction reversal (Simulus), and
  action-response failure (STORM). Failure types are representative rather than
  model-specific; the same failure may occur in multiple world models.}
  \label{fig:closed-loop-rollout-failures}
\ifdefined\WACVVERSION
\end{figure*}
\else
\end{figure}
\fi

\subsection{Native Zero-Shot MBRL}
\label{sec:native-zero-shot-mbrl}

We next evaluate whether each frozen world model can support standalone policy
learning under zero-shot MBRL. Because policy performance depends on both the
world model and RL method, we preserve each reproduced agent's \emph{native}
policy-learning setup while freezing its final model. We call this protocol
\emph{native zero-shot MBRL}.

Native policies operate on generated pixels (DIAMOND), learned latent states
(DreamerV3, TWISTER, and STORM), or latent tokens (Simulus). The corresponding
architectures and prediction spaces are summarized in the supplementary
material.

The results use one training seed per reproduced agent. Each policy is evaluated
over 20 real-ALE Pong episodes with deterministic action selection, and
\autoref{tab:native-zero-shot-mbrl-in-pong} reports the mean $\pm$ sample
standard deviation across episodes.

Using this native protocol, policies trained inside the frozen WMs
(\emph{Frozen-WM policies}) substantially underperform the corresponding
reproduced policies (\emph{Dyna-style policies}) across all five projects
(\autoref{fig:main-results}(a),
\autoref{tab:native-zero-shot-mbrl-in-pong}).
The degradation is severe: in four of the five models, the return drops by
more than 29 points.

This agent--world-model gap, together with the ball-related rollout failures
observed in \autoref{sec:reproduced-wm-closed-loop-rollout-diagnosis},
suggests that task-critical concepts, particularly the Pong
ball, may not be sufficiently learned. This motivates \methodname{},
introduced in the next section.

\FloatBarrier
\begin{table}[htbp]
  \centering
  \scriptsize
  \setlength{\tabcolsep}{2.1pt}
  \caption{
  \textbf{Agent--WM gap under native zero-shot MBRL in Pong.}
  Our scores use one training seed per reproduced agent and report the mean
  $\pm$ sample standard deviation across 20 real-ALE evaluation episodes.
  Published scores are taken from the corresponding papers and may use
  different evaluation protocols; they are included only as reproduction
  references.
  $\Delta$ denotes Frozen-WM policy minus reproduced Dyna-style policy.
  Bold marks the higher score between the two; higher is better.
  }
  \label{tab:native-zero-shot-mbrl-in-pong}

  \begin{tabular*}{\columnwidth}{@{\extracolsep{\fill}}lrrrr@{}}
    \toprule
    Model &
    Pub. &
    Dyna policy &
    Frozen policy &
    $\Delta$ \\
    \midrule

    DreamerV3 &
    $-4.0$ &
    $\mathbf{-5.5 \pm 3.3}$ &
    $-20.9 \pm 0.3$ &
    $-15.5$ \\

    DIAMOND &
    $20.4$ &
    $\mathbf{19.7 \pm 2.4}$ &
    $-9.6 \pm 12.6$ &
    $-29.3$ \\

    TWISTER &
    $20.0$ &
    $\mathbf{17.7 \pm 1.3}$ &
    $-13.3 \pm 7.8$ &
    $-31.0$ \\

    Simulus &
    $19.9$ &
    $\mathbf{20.8 \pm 0.6}$ &
    $-11.6 \pm 8.3$ &
    $-32.4$ \\

    STORM &
    $11.0$ &
    $\mathbf{18.7 \pm 1.6}$ &
    $-21.0 \pm 0.0$ &
    $-39.7$ \\

    \bottomrule
  \end{tabular*}
\end{table}

\paragraph{Beyond Pong.}

The gap is not limited to Pong. Across the Atari100K
benchmark~\citep{kaiser2020simple}, Frozen-WM policies score lower than their
corresponding Dyna-style policies in 23 of 26 games for DreamerV3, 22 of 26
for TWISTER, 22 of 26 for Simulus, and all 26 games for STORM
(\autoref{fig:atari100k-frozen-wm-summary}).

To compare games with different score scales, we normalize the mean-return gap
by each game's human--random score range:
\[
\Delta_{\mathrm{HNS}}
=
100\frac{S_{\mathrm{Frozen}}-S_{\mathrm{Reproduced}}}
{S_{\mathrm{Human}}-S_{\mathrm{Random}}} .
\]

The normalized gap is encoded by color, with blue and orange indicating lower
and higher Frozen-WM performance, respectively, and color intensity indicating
the magnitude.

Exact ties occur only in Freeway and Private Eye. Freeway has an extremely
sparse reward structure: the agent must repeatedly move upward through traffic
to complete a crossing before receiving any reward, and the tied policies fail
to reach this first reward.
Private Eye is exploration-heavy and spans multiple scenes; the tied policies
obtain only the small reward available in the initial scene and fail to make
further progress.
Detailed
per-game results are provided in the supplementary material.
\section{Concept-Guided Spatial Regularization}
\label{sec:cgsreg}

\methodname{} adds extra reconstruction supervision to image regions
corresponding to task-critical concepts during world model training. Modern visual world models, including
the five studied here, typically combine an image reconstruction or prediction
loss with other objectives such as dynamics, reward, and continuation
prediction. We therefore write
\begin{equation}
L_{\mathrm{wm}}
=
L_{\mathrm{img}}
+
L_{\mathrm{nonvisual}},
\label{eq:base-wm-objective}
\end{equation}
where $L_{\mathrm{img}}$ is the image loss and
$L_{\mathrm{nonvisual}}$ collects the remaining world-model objectives.

We introduce \methodname{} as
\begin{equation}
L_{\mathrm{wm}}
=
L_{\mathrm{img}}
+
\lambda_{\mathrm{CGSReg}}L_{\mathrm{CGSReg}}
+
L_{\mathrm{nonvisual}}.
\label{eq:cgsreg-objective}
\end{equation}

Here, $\lambda_{\mathrm{CGSReg}} \ge 0$ controls the strength of the
concept-region regularization, with $\lambda_{\mathrm{CGSReg}}=0$ recovering
the original world-model objective.

For a target image $x$, prediction
$\hat{x}$, and binary concept mask $m$ broadcast over image channels, we define

\begin{equation}
L_{\mathrm{CGSReg}}
=
\frac{
\sum_p m_p(x_p-\hat{x}_p)^2
}{
\max\left(1,\sum_p m_p\right)
} .
\label{eq:cgsreg-loss}
\end{equation}
where $p$ indexes image values. The normalization makes the loss depend on the
average reconstruction error within the concept region rather than its spatial
size, while the safe denominator $\max\left(1,\sum_p m_p\right)$ handles empty masks.

For Pong, we use the ball as the task-critical concept and obtain its mask
from replay observations with SAM2~\citep{ravi2024sam2}. The masks are used only during world-model training.

\section{Experiments}
\label{sec:experiments}

We evaluate \methodname{} using closed-loop rollout diagnosis and unified
pixel-space zero-shot MBRL.

\subsection{Offline World-Model Training}
\label{sec:offline-world-model-training}

To isolate world-model training from the data-collection and
policy-learning loops of the original Dyna-style agents, we train on the same fixed 100k-step Pong replay
collected from our DIAMOND reproduction, without additional environment
interaction or policy updates. We refer to this setting as
\emph{offline world-model training}. For each architecture, we train two
configurations, each with three world-model training seeds:

\begin{itemize}[leftmargin=*]
  \item \textbf{Offline baseline WM}: trained with
  $\lambda_{\mathrm{CGSReg}}=0$.

  \item \textbf{Offline \methodname{} WM}: trained with a selected nonzero
  $\lambda_{\mathrm{CGSReg}}$.
\end{itemize}

Within each architecture, the two configurations use the same world-model
update budget. We evaluate each seed with the same 20-episode real-ALE protocol,
average episode returns within each seed, and report the mean $\pm$ sample
standard deviation across the three world-model seeds.

For each architecture, we tune $\lambda_{\mathrm{CGSReg}}$ on the same DIAMOND
replay by evaluating $\{0.01,0.1,1.0\}$ with unified pixel-space zero-shot MBRL
and selecting the highest mean return across three world-model seeds. The
selected values are $0.1$ for DreamerV3, $0.01$ for DIAMOND, $1.0$ for TWISTER,
and $0.01$ for Simulus and STORM; all STORM candidates perform equally.

Offline training details and the full regularization-weight and replay-dataset
ablations are provided in the supplementary material.

\subsection{Closed-Loop Rollout Diagnosis}
\label{sec:cgsreg-closed-loop-rollout-diagnosis}

\begin{figure*}[!t]
  \centering
  \includegraphics[width=0.96\textwidth]{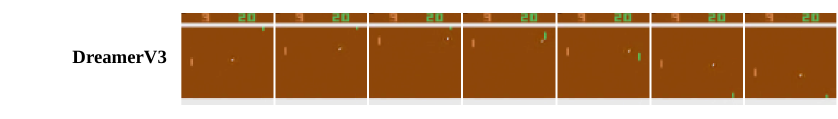}
  \caption{\textbf{\methodname{}-improved DreamerV3 closed-loop rollout.}
  The frames are sampled from a horizon-512 rollout driven by the same external
  Pong controller as in
  \autoref{fig:closed-loop-rollout-failures}. The ball remains visible,
  rebounds from the player paddle, and continues along a plausible trajectory.}
  \label{fig:cgsreg-dreamerv3-closed-loop-rollout}
\end{figure*}

We apply the closed-loop diagnostic from
\autoref{sec:reproduced-wm-closed-loop-rollout-diagnosis} to the offline baseline and \methodname{}
world models. For clarity, \autoref{fig:cgsreg-dreamerv3-closed-loop-rollout} shows only DreamerV3, where the qualitative improvement is the clearest.
For the other models, DIAMOND and TWISTER also show improved ball modeling and dynamics, although occasional errors remain. Simulus shows no clear rollout improvement, while STORM continues to exhibit unreliable action response.

\subsection{Unified Pixel-Space Zero-Shot MBRL}
\label{sec:unified-pixel-space-zero-shot-mbrl}

We next evaluate \methodname{} with \emph{pixel-space zero-shot MBRL}, where
the policy learns directly from pixel observations generated by the frozen
world model. To control for differences in native RL methods, we train the same
image-based on-policy actor--critic from scratch for 20k updates, with identical
architecture, learning procedure, and hyperparameters across all checkpoints.
We call this \emph{unified pixel-space zero-shot MBRL}.

\autoref{fig:main-results}(b) summarizes the controlled comparison.
Compared with the matched offline baselines, \methodname{} improves mean
real-environment Pong return by $5.8$, $8.6$, $12.0$, and $11.7$ points
for DreamerV3, DIAMOND, TWISTER, and Simulus, respectively
(\autoref{tab:unified-pixel-space-zero-shot-mbrl-in-pong}).
STORM remains at the minimum return of $-21$.

The reproduced-agent WMs are included as an additional reference in
\autoref{tab:unified-pixel-space-zero-shot-mbrl-in-pong}. Under the unified protocol,
offline \methodname{} WMs outperform reproduced-agent WMs for DreamerV3,
DIAMOND, TWISTER, and Simulus, although all evaluated WMs remain far from 21.

\begin{table}[htbp]
  \centering
  \scriptsize
  \setlength{\tabcolsep}{2.1pt}
  \caption{
    \textbf{Effect of \methodname{} under unified pixel-space zero-shot MBRL in Pong.}
    Matched offline baseline and \methodname{} WMs form the main comparison;
    reproduced-agent WMs are references. Offline entries report the mean
    $\pm$ sample standard deviation over three WM seeds, each evaluated on 20
    real-ALE episodes; reproduced-agent WMs use one seed and the same evaluation.
    Bold marks the better offline result; higher is better.
  }

  \label{tab:unified-pixel-space-zero-shot-mbrl-in-pong}

  \begin{tabular*}{\columnwidth}{@{\extracolsep{\fill}}lrrr@{}}

    \toprule

    Model &
    Reproduced &
    Baseline &
    \methodname{} \\

    \midrule

    DreamerV3 &
    $-21.0 \pm 0.2$ &
    $-19.2 \pm 3.2$ &
    $\mathbf{-13.4 \pm 1.7}$ \\

    DIAMOND &
    $-19.7 \pm 1.5$ &
    $-9.6 \pm 4.5$ &
    $\mathbf{-1.0 \pm 4.2}$ \\

    TWISTER &
    $-8.4 \pm 6.4$ &
    $-20.3 \pm 0.9$ &
    $\mathbf{-8.3 \pm 10.5}$ \\

    Simulus &
    $-9.3 \pm 9.6$ &
    $-16.8 \pm 3.8$ &
    $\mathbf{-5.1 \pm 5.7}$ \\

    STORM &
    $-21.0 \pm 0.0$ &
    $-21.0 \pm 0.0$ &
    $-21.0 \pm 0.0$ \\

    \bottomrule

  \end{tabular*}
\end{table}

\subsection{Outcomes}
\label{sec:experiment-outcomes}

The effects of \methodname{} differ across architectures. Simulus uses a two-stage pipeline that first trains an image tokenizer and then a separate token-dynamics model. Because \methodname{} is a visual reconstruction loss, we apply it only during image-tokenizer pretraining, so it does not supervise the subsequent dynamics model. Its pixel-space zero-shot MBRL performance improves, but we observe no clear rollout improvement. In STORM, the dominant failure is poor action response, which ball-region reconstruction does not directly address; neither diagnostic improves clearly.

Overall, \methodname{} improves both diagnostics in DreamerV3, DIAMOND, and
TWISTER, indicating more reliable visual and dynamical trajectory generation
and stronger use of the frozen world model as an RL simulator. Taken together, these results show that \methodname{} can broadly improve
frozen visual world models, although its effect may be limited when
other modeling bottlenecks dominate, as in Simulus and STORM.

\section{Limitations and Discussion}
\label{sec:limitations}

\paragraph{\methodname{} is not sufficient.}
\methodname{} improves pixel-space zero-shot MBRL in four architectures and
closed-loop rollouts in three, but all resulting policies remain far from the
maximum Pong return of $21$
(\autoref{tab:unified-pixel-space-zero-shot-mbrl-in-pong}).
Ball-region regularization therefore addresses only part of this
gap.

\paragraph{The scope is limited.}
We evaluate \methodname{} only in Pong and on five world-model architectures
ranging from roughly 10M to 220M parameters. Its generalization to broader tasks
and larger models remains open.

\paragraph{The concepts are manually specified.}
\methodname{} assumes that task-relevant concepts are known and can be mapped to
pixel regions. This works well for localized concepts such as the Pong ball, but
does not naturally extend to world rules or physical mechanisms that are not
localized in pixel space.
In the spirit of \emph{The Bitter Lesson}~\citep{sutton2019bitter}, more general methods should leverage
increased computation to discover concepts
automatically and reduce reliance on human priors~\citep{sutton2009predictive,sutton2011horde}.

\paragraph{Why does the agent--world-model gap exist?}

One possible explanation is that Dyna-style training does not require the world
model to be reliable. Instead, the model may act primarily as a local
data augmentor, generating trajectories near the evolving policy
distribution while continually fitting newly collected real experience.
It can therefore support policy improvement without serving as a reliable
standalone simulator. This makes the learning process closer in spirit
to model-free RL, which is known to perform well on Atari.

We test this hypothesis by freezing the world model at different training stages
while continuing policy learning, preventing it from adapting to newly collected
data. The results provide partial support: DreamerV3 and DIAMOND degrade after
freezing, TWISTER shows a weaker effect, Simulus does not degrade, and STORM is
non-monotonic. Thus, continual world-model updating explains part, but not all,
of the agent--world-model gap.

\section{Conclusion}
\label{sec:conclusion}

We identify a widespread agent--world-model gap across Pong and Atari100K:
strong Dyna-style agents can contain weak standalone world models. In Pong,
motivated by visual trajectory failures, we propose \methodname{}, which improves
frozen world models across several architectures.

\section*{Acknowledgments}

We thank Massed Compute for providing computational resources.

\FloatBarrier

{
  \small
  \bibliographystyle{ieeenat_fullname}
  \bibliography{references}
}

\clearpage
\onecolumn
\appendix
\makeatletter
\@addtoreset{algorithm}{section}
\makeatother
\renewcommand{\thealgorithm}{\thesection.\arabic{algorithm}}
\providecommand{\theHalgorithm}{}
\renewcommand{\theHalgorithm}{appendix.\thesection.\arabic{algorithm}}

\section{Frozen-WM Evaluation across Atari100K}
\label{app:atari100k-frozen-wm-diagnostics}

We extend the frozen-WM evaluation beyond Pong to all 26 Atari100K
games using native zero-shot MBRL
(\mainNativeZeroShotRef). For each evaluated project, we first
reproduce the Dyna-style agent and then train a new policy inside its final
frozen world model. This section provides the evaluation protocol,
human-normalization references used in
\mainNativeZeroShotRef, and detailed per-game results.

\subsection{Evaluation Protocol and Compute}
\label{app:atari100k-evaluation-protocol}

Published scores, where reported, are taken from the corresponding world-model
agent papers and are included only as references for reproduction quality;
they may use evaluation protocols different from ours.

Each row uses one RL training seed per policy condition and reports the
mean $\pm$ sample standard deviation across 20 real-ALE evaluation episodes
with reset seeds 0--19.

There are two protocol exceptions. For Simulus non-Pong games, we use random
no-op starts with \texttt{noop\_max=30} to avoid repeated trajectories. For
Breakout, the action sampling strategy used in evaluation is stochastic rather than deterministic.

\begin{table}[htbp]
  \centering
  \small
  \setlength{\tabcolsep}{6pt}
  \renewcommand{\arraystretch}{1.12}
  \caption{\textbf{Per-game compute cost for the Atari100K extension.}
  Agent reproduction and native zero-shot MBRL are run sequentially on a
  single RTX 3090.}
  \label{tab:atari100k-cross-task-runtime}

  \begin{tabular}{@{}lccc@{}}
    \toprule
    Project & Agent reproduction & Native ZS MBRL & Total \\
    \midrule
    DreamerV3 & $4$--$4.5$ h   & $2.4$--$2.6$ h & $6.5$--$7$ h \\
    DIAMOND   & $145$--$150$ h & $260$--$280$ h & $16.9$--$17.9$ d \\
    TWISTER   & $10$--$11$ h   & $7$--$8$ h     & $17$--$19$ h \\
    Simulus   & $23$--$25$ h   & $9.3$--$9.6$ h & $32.3$--$34.6$ h \\
    STORM     & $5$--$5.5$ h   & $2.7$--$3.2$ h & $7.7$--$8.7$ h \\
    \bottomrule
  \end{tabular}
\end{table}

We omit the full Atari100K evaluation for DIAMOND due to its high
computational cost.

\FloatBarrier
\subsection{Human-Normalization References}
\label{app:human-normalization-references}

For the human-normalized mean-return gap defined in the main paper, we use the
random-policy and professional-human scores reported by
\citet{mnih2015human} as fixed game-specific references. The same reference
values are distributed with DeepMind's DQN Zoo.
These values are not estimated from our evaluation runs. The table below lists
the random and human reference scores used for all 26 Atari100K games.

\begin{table*}[t]
  \centering
  \scriptsize
  \caption{
    \textbf{Atari reference scores used for human normalization.}
    Random and Human are the canonical random-policy and
    professional-human mean returns from \citet{mnih2015human}.
    These fixed values are used only to normalize the return gaps in
    \mainAtariGapFigureRef.
  }
  \setlength{\tabcolsep}{4.5pt}
  \begin{tabular}{lrr@{\hspace{12pt}}lrr}
    \toprule
    Game & Random & Human &
    Game & Random & Human \\
    \midrule
    Alien           & 227.8   & 7127.7  &
    Gopher          & 257.6   & 2412.5  \\
    Amidar          & 5.8     & 1719.5  &
    Hero            & 1027.0  & 30826.4 \\
    Assault         & 222.4   & 742.0   &
    James Bond      & 29.0    & 302.8   \\
    Asterix         & 210.0   & 8503.3  &
    Kangaroo        & 52.0    & 3035.0  \\
    Bank Heist      & 14.2    & 753.1   &
    Krull           & 1598.0  & 2665.5  \\
    Battle Zone     & 2360.0  & 37187.5 &
    Kung-Fu Master  & 258.5   & 22736.3 \\
    Boxing          & 0.1     & 12.1    &
    Ms.\ Pac-Man    & 307.3   & 6951.6  \\
    Breakout        & 1.7     & 30.5    &
    Pong            & -20.7   & 14.6    \\
    Chopper Command & 811.0   & 7387.8  &
    Private Eye     & 24.9    & 69571.3 \\
    Crazy Climber   & 10780.5 & 35829.4 &
    Qbert           & 163.9   & 13455.0 \\
    Demon Attack    & 152.1   & 1971.0  &
    Road Runner     & 11.5    & 7845.0  \\
    Freeway         & 0.0     & 29.6    &
    Seaquest        & 68.4    & 42054.7 \\
    Frostbite       & 65.2    & 4334.7  &
    Up N Down       & 533.4   & 11693.2 \\
    \bottomrule
  \end{tabular}
\end{table*}

\FloatBarrier
\subsection{Per-Game Results}
\label{app:atari100k-per-game-results}

Tables below report the complete per-game results. Bold indicates the higher
score between the reproduced agent and Frozen-WM policy; ties are not bolded.
Higher is better, with
\[
\Delta
=
\text{Frozen-WM policy}
-
\text{Reproduced agent}.
\]

\begingroup

\makeatletter
\setlength{\@fptop}{0pt}
\setlength{\@fpbot}{0pt plus 1fil}
\makeatother

\clearpage
\begin{table}[p]
  \centering
  \footnotesize
  \setlength{\tabcolsep}{3.0pt}
  \renewcommand{\arraystretch}{1.14}
  \caption{\textbf{DreamerV3 frozen-WM diagnostic across Atari100K.}
  The frozen-WM policy scores lower than the reproduced agent in 23 of 26
  games, ties in two (Freeway and Private Eye), and scores higher in one (Krull).}
  \label{tab:dreamerv3-atari100k-frozen-wm}

  \begin{tabular}{@{}lrrrr@{}}
    \toprule
    Atari task & Published & Reproduced agent & Frozen-WM policy & $\Delta$ \\
    \midrule
    Alien &
    $1{,}118.0$ &
    $\mathbf{885.0 \pm 446.8}$ &
    $265.5 \pm 160.2$ &
    $-619.5$ \\

    Amidar &
    $97.0$ &
    $\mathbf{293.6 \pm 16.7}$ &
    $46.8 \pm 27.7$ &
    $-246.9$ \\

    Assault &
    $683.0$ &
    $\mathbf{606.9 \pm 125.1}$ &
    $427.4 \pm 117.9$ &
    $-179.6$ \\

    Asterix &
    $1{,}062.0$ &
    $\mathbf{970.0 \pm 373.6}$ &
    $515.0 \pm 208.4$ &
    $-455.0$ \\

    Bank Heist &
    $398.0$ &
    $\mathbf{145.5 \pm 23.5}$ &
    $96.0 \pm 27.8$ &
    $-49.5$ \\

    Battle Zone &
    $20{,}300.0$ &
    $\mathbf{15{,}000.0 \pm 3{,}128.8}$ &
    $5{,}300.0 \pm 3{,}341.9$ &
    $-9{,}700.0$ \\

    Boxing &
    $82.0$ &
    $\mathbf{82.7 \pm 6.9}$ &
    $-0.7 \pm 5.3$ &
    $-83.4$ \\

    Breakout &
    $10.0$ &
    $\mathbf{8.6 \pm 3.4}$ &
    $7.4 \pm 5.6$ &
    $-1.3$ \\

    Chopper Command &
    $2{,}222.0$ &
    $\mathbf{1{,}380.0 \pm 359.2}$ &
    $685.0 \pm 218.3$ &
    $-695.0$ \\

    Crazy Climber &
    $86{,}225.0$ &
    $\mathbf{47{,}130.0 \pm 15{,}399.6}$ &
    $17{,}565.0 \pm 13{,}389.8$ &
    $-29{,}565.0$ \\

    Demon Attack &
    $577.0$ &
    $\mathbf{236.0 \pm 113.0}$ &
    $82.3 \pm 45.8$ &
    $-153.8$ \\

    Freeway &
    $0.0$ &
    $0.0 \pm 0.0$ &
    $0.0 \pm 0.0$ &
    $0.0$ \\

    Frostbite &
    $3{,}377.0$ &
    $\mathbf{250.5 \pm 21.4}$ &
    $216.5 \pm 12.3$ &
    $-34.0$ \\

    Gopher &
    $2{,}160.0$ &
    $\mathbf{1{,}101.0 \pm 472.2}$ &
    $212.0 \pm 127.9$ &
    $-889.0$ \\

    Hero &
    $13{,}354.0$ &
    $\mathbf{13{,}370.3 \pm 32.5}$ &
    $2{,}967.5 \pm 55.1$ &
    $-10{,}402.8$ \\

    James Bond &
    $540.0$ &
    $\mathbf{222.5 \pm 88.1}$ &
    $32.5 \pm 40.6$ &
    $-190.0$ \\

    Kangaroo &
    $2{,}643.0$ &
    $\mathbf{1{,}680.0 \pm 313.9}$ &
    $770.0 \pm 357.0$ &
    $-910.0$ \\

    Krull &
    $8{,}171.0$ &
    $6{,}122.0 \pm 1{,}225.5$ &
    $\mathbf{6{,}662.0 \pm 1{,}209.3}$ &
    $+540.0$ \\

    Kung Fu Master &
    $25{,}900.0$ &
    $\mathbf{20{,}649.0 \pm 6{,}585.8}$ &
    $7{,}590.0 \pm 1{,}981.5$ &
    $-13{,}059.0$ \\

    Ms. Pac-Man &
    $1{,}521.0$ &
    $\mathbf{2{,}953.5 \pm 333.8}$ &
    $419.0 \pm 162.4$ &
    $-2{,}534.5$ \\

    Pong &
    $-4.0$ &
    $\mathbf{-5.5 \pm 3.3}$ &
    $-20.9 \pm 0.3$ &
    $-15.5$ \\

    Private Eye &
    $3{,}238.0$ &
    $100.0 \pm 0.0$ &
    $100.0 \pm 0.0$ &
    $0.0$ \\

    Qbert &
    $2{,}921.0$ &
    $\mathbf{4{,}552.5 \pm 766.4}$ &
    $433.8 \pm 78.8$ &
    $-4{,}118.8$ \\

    Road Runner &
    $19{,}230.0$ &
    $\mathbf{12{,}790.0 \pm 4{,}334.1}$ &
    $1{,}645.0 \pm 750.1$ &
    $-11{,}145.0$ \\

    Seaquest &
    $962.0$ &
    $\mathbf{412.0 \pm 112.8}$ &
    $124.0 \pm 28.0$ &
    $-288.0$ \\

    Up N Down &
    $46{,}910.0$ &
    $\mathbf{9{,}050.5 \pm 2{,}047.4}$ &
    $3{,}523.5 \pm 891.9$ &
    $-5{,}527.0$ \\
    \bottomrule
  \end{tabular}
\end{table}

\clearpage
\begin{table}[p]
  \centering
  \footnotesize
  \setlength{\tabcolsep}{3.0pt}
  \renewcommand{\arraystretch}{1.14}
  \caption{\textbf{TWISTER frozen-WM diagnostic across Atari100K.}
  The frozen-WM policy scores lower than the reproduced agent in 22 of 26
  games, ties in one (Freeway), and scores higher in three (Assault, Chopper
  Command and Private Eye).}
  \label{tab:twister-atari100k-frozen-wm}

  \begin{tabular}{@{}lrrrr@{}}
    \toprule
    Atari task & Published & Reproduced agent & Frozen-WM policy & $\Delta$ \\
    \midrule
    Alien &
    $970.0$ &
    $\mathbf{1{,}066.5 \pm 372.9}$ &
    $357.0 \pm 174.8$ &
    $-709.5$ \\

    Amidar &
    $184.0$ &
    $\mathbf{218.5 \pm 33.7}$ &
    $30.8 \pm 24.1$ &
    $-187.7$ \\

    Assault &
    $721.0$ &
    $802.2 \pm 150.8$ &
    $\mathbf{866.3 \pm 458.8}$ &
    $+64.1$ \\

    Asterix &
    $1{,}306.0$ &
    $\mathbf{1{,}592.5 \pm 422.8}$ &
    $320.0 \pm 116.3$ &
    $-1{,}272.5$ \\

    Bank Heist &
    $942.0$ &
    $\mathbf{1{,}222.5 \pm 45.8}$ &
    $1{,}163.5 \pm 69.2$ &
    $-59.0$ \\

    Battle Zone &
    $9{,}920.0$ &
    $\mathbf{9{,}300.0 \pm 3{,}948.4}$ &
    $2{,}350.0 \pm 2{,}641.3$ &
    $-6{,}950.0$ \\

    Boxing &
    $88.0$ &
    $\mathbf{97.8 \pm 3.6}$ &
    $-0.7 \pm 4.9$ &
    $-98.5$ \\

    Breakout &
    $35.0$ &
    $\mathbf{18.1 \pm 10.7}$ &
    $5.2 \pm 3.1$ &
    $-13.0$ \\

    Chopper Command &
    $910.0$ &
    $845.0 \pm 287.4$ &
    $\mathbf{1{,}035.0 \pm 169.4}$ &
    $+190.0$ \\

    Crazy Climber &
    $81{,}880.0$ &
    $\mathbf{87{,}665.0 \pm 33{,}322.2}$ &
    $81{,}410.0 \pm 32{,}284.3$ &
    $-6{,}255.0$ \\

    Demon Attack &
    $289.0$ &
    $\mathbf{444.0 \pm 363.7}$ &
    $93.3 \pm 68.8$ &
    $-350.8$ \\

    Freeway &
    $32.0$ &
    $0.0 \pm 0.0$ &
    $0.0 \pm 0.0$ &
    $0.0$ \\

    Frostbite &
    $305.0$ &
    $\mathbf{277.0 \pm 22.0}$ &
    $107.0 \pm 27.7$ &
    $-170.0$ \\

    Gopher &
    $22{,}234.0$ &
    $\mathbf{1{,}817.0 \pm 1{,}299.0}$ &
    $515.0 \pm 411.2$ &
    $-1{,}302.0$ \\

    Hero &
    $8{,}773.0$ &
    $\mathbf{7{,}612.5 \pm 52.5}$ &
    $7{,}474.0 \pm 73.3$ &
    $-138.5$ \\

    James Bond &
    $573.0$ &
    $\mathbf{670.0 \pm 153.4}$ &
    $107.5 \pm 52.0$ &
    $-562.5$ \\

    Kangaroo &
    $6{,}016.0$ &
    $\mathbf{6{,}765.0 \pm 1{,}839.4}$ &
    $120.0 \pm 119.6$ &
    $-6{,}645.0$ \\

    Krull &
    $8{,}839.0$ &
    $\mathbf{8{,}581.0 \pm 1{,}556.8}$ &
    $7{,}823.5 \pm 1{,}561.4$ &
    $-757.5$ \\

    Kung Fu Master &
    $23{,}442.0$ &
    $\mathbf{27{,}465.0 \pm 9{,}725.1}$ &
    $20{,}610.0 \pm 7{,}090.5$ &
    $-6{,}855.0$ \\

    Ms. Pac-Man &
    $2{,}206.0$ &
    $\mathbf{2{,}350.5 \pm 587.9}$ &
    $979.0 \pm 404.6$ &
    $-1{,}371.5$ \\

    Pong &
    $20.0$ &
    $\mathbf{17.7 \pm 1.3}$ &
    $-13.3 \pm 7.8$ &
    $-31.0$ \\

    Private Eye &
    $1{,}608.0$ &
    $78.6 \pm 95.9$ &
    $\mathbf{90.0 \pm 30.8}$ &
    $+11.5$ \\

    Qbert &
    $3{,}197.0$ &
    $\mathbf{3{,}873.8 \pm 1{,}283.4}$ &
    $373.8 \pm 86.8$ &
    $-3{,}500.0$ \\

    Road Runner &
    $17{,}832.0$ &
    $\mathbf{19{,}940.0 \pm 2{,}468.2}$ &
    $2{,}490.0 \pm 1{,}690.2$ &
    $-17{,}450.0$ \\

    Seaquest &
    $532.0$ &
    $\mathbf{615.0 \pm 161.4}$ &
    $28.0 \pm 17.7$ &
    $-587.0$ \\

    Up N Down &
    $7{,}068.0$ &
    $\mathbf{43{,}675.0 \pm 29{,}679.7}$ &
    $722.0 \pm 484.6$ &
    $-42{,}953.0$ \\
    \bottomrule
  \end{tabular}
\end{table}

\clearpage
\begin{table}[p]
  \centering
  \footnotesize
  \setlength{\tabcolsep}{3.0pt}
  \renewcommand{\arraystretch}{1.14}
  \caption{\textbf{Simulus frozen-WM diagnostic across Atari100K.}
  The frozen-WM policy scores lower than the reproduced agent in 22 of 26
  games, ties in one (Private Eye), and scores higher in three (Demon Attack,
  Hero, and Krull).
  Values are raw-return mean $\pm$ sample standard deviation across 20
  real-ALE evaluation episodes.}
  \label{tab:simulus-atari100k-frozen-wm}

  \begin{tabular}{@{}lrrrr@{}}
    \toprule
    Atari task & Published & Reproduced agent & Frozen-WM policy & $\Delta$ \\
    \midrule
    Alien & $687.2$ & $\mathbf{606.0 \pm 15.7}$ & $272.0 \pm 41.0$ & $-334.0$ \\
    Amidar & $102.4$ & $\mathbf{111.8 \pm 25.0}$ & $81.6 \pm 4.0$ & $-30.2$ \\
    Assault & $1{,}822.8$ & $\mathbf{2{,}126.3 \pm 903.5}$ & $1{,}692.6 \pm 947.2$ & $-433.7$ \\
    Asterix & $1{,}369.1$ & $\mathbf{1{,}302.5 \pm 329.9}$ & $1{,}150.0 \pm 433.2$ & $-152.5$ \\
    Bank Heist & $347.1$ & $\mathbf{333.0 \pm 15.6}$ & $233.5 \pm 5.9$ & $-99.5$ \\
    Battle Zone & $13{,}262.0$ & $\mathbf{10{,}850.0 \pm 3{,}745.5}$ & $2{,}400.0 \pm 502.6$ & $-8{,}450.0$ \\
    Boxing & $93.5$ & $\mathbf{97.0 \pm 3.6}$ & $67.1 \pm 10.8$ & $-29.9$ \\
    Breakout & $148.9$ & $\mathbf{161.5 \pm 89.3}$ & $138.3 \pm 105.3$ & $-23.2$ \\
    Chopper Command & $3{,}611.6$ & $\mathbf{3{,}310.0 \pm 236.0}$ & $1{,}885.0 \pm 798.9$ & $-1{,}425.0$ \\
    Crazy Climber & $93{,}433.2$ & $\mathbf{92{,}185.0 \pm 4{,}692.8}$ & $73{,}055.0 \pm 15{,}597.1$ & $-19{,}130.0$ \\
    Demon Attack & $4{,}787.6$ & $5{,}896.3 \pm 1{,}467.0$ & $\mathbf{6{,}719.5 \pm 956.4}$ & $+823.3$ \\
    Freeway & $31.9$ & $\mathbf{31.6 \pm 0.5}$ & $30.8 \pm 0.6$ & $-0.8$ \\
    Frostbite & $258.4$ & $\mathbf{272.0 \pm 4.1}$ & $200.0 \pm 0.0$ & $-72.0$ \\
    Gopher & $4{,}363.2$ & $\mathbf{4{,}354.0 \pm 1{,}383.6}$ & $837.0 \pm 400.6$ & $-3{,}517.0$ \\
    Hero & $7{,}466.8$ & $7{,}425.0 \pm 0.0$ & $\mathbf{7{,}725.0 \pm 0.0}$ & $+300.0$ \\
    James Bond & $678.0$ & $\mathbf{605.0 \pm 128.7}$ & $422.5 \pm 195.7$ & $-182.5$ \\
    Kangaroo & $6{,}656.0$ & $\mathbf{8{,}220.0 \pm 1{,}428.9}$ & $0.0 \pm 0.0$ & $-8{,}220.0$ \\
    Krull & $6{,}677.3$ & $6{,}142.5 \pm 630.7$ & $\mathbf{7{,}443.5 \pm 1{,}137.8}$ & $+1{,}301.0$ \\
    Kung Fu Master & $31{,}705.4$ & $\mathbf{36{,}240.0 \pm 7{,}536.5}$ & $22{,}155.0 \pm 3{,}036.4$ & $-14{,}085.0$ \\
    Ms. Pac-Man & $1{,}282.7$ & $\mathbf{947.0 \pm 279.6}$ & $620.0 \pm 299.7$ & $-327.0$ \\
    Pong & $19.9$ & $\mathbf{20.8 \pm 0.6}$ & $-11.6 \pm 8.3$ & $-32.4$ \\
    Private Eye & $100.0$ & $100.0 \pm 0.0$ & $100.0 \pm 0.0$ & $0.0$ \\
    Qbert & $2{,}425.6$ & $\mathbf{875.0 \pm 51.3}$ & $835.0 \pm 82.1$ & $-40.0$ \\
    Road Runner & $24{,}471.8$ & $\mathbf{33{,}255.0 \pm 7{,}028.2}$ & $27{,}545.0 \pm 7{,}491.1$ & $-5{,}710.0$ \\
    Seaquest & $1{,}800.4$ & $\mathbf{1{,}750.0 \pm 133.8}$ & $147.0 \pm 57.8$ & $-1{,}603.0$ \\
    Up N Down & $10{,}416.5$ & $\mathbf{4{,}383.5 \pm 1{,}076.0}$ & $3{,}768.5 \pm 810.8$ & $-615.0$ \\
    \bottomrule
  \end{tabular}
\end{table}

\clearpage
\begin{table}[p]
  \centering
  \footnotesize
  \setlength{\tabcolsep}{3.0pt}
  \renewcommand{\arraystretch}{1.14}
  \caption{\textbf{STORM frozen-WM diagnostic across Atari100K.}
  The frozen-WM policy scores lower than the reproduced agent in all 26 games.}
  \label{tab:storm-atari100k-frozen-wm}

  \begin{tabular}{@{}lrrrr@{}}
    \toprule
    Atari task & Published & Reproduced agent & Frozen-WM policy & $\Delta$ \\
    \midrule
    Alien &
    $984.0$ &
    $\mathbf{522.5 \pm 177.7}$ &
    $55.0 \pm 8.9$ &
    $-467.5$ \\

    Amidar &
    $205.0$ &
    $\mathbf{92.6 \pm 32.0}$ &
    $0.0 \pm 0.0$ &
    $-92.6$ \\

    Assault &
    $801.0$ &
    $\mathbf{897.0 \pm 166.0}$ &
    $0.0 \pm 0.0$ &
    $-897.0$ \\

    Asterix &
    $1{,}028.0$ &
    $\mathbf{967.5 \pm 198.9}$ &
    $155.0 \pm 70.5$ &
    $-812.5$ \\

    Bank Heist &
    $641.0$ &
    $\mathbf{852.5 \pm 94.9}$ &
    $0.0 \pm 0.0$ &
    $-852.5$ \\

    Battle Zone &
    $13{,}540.0$ &
    $\mathbf{3{,}250.0 \pm 2{,}446.8}$ &
    $2{,}400.0 \pm 1{,}635.1$ &
    $-850.0$ \\

    Boxing &
    $80.0$ &
    $\mathbf{74.3 \pm 13.0}$ &
    $-12.4 \pm 7.1$ &
    $-86.6$ \\

    Breakout &
    $16.0$ &
    $\mathbf{36.9 \pm 69.4}$ &
    $0.0 \pm 0.0$ &
    $-36.9$ \\

    Chopper Command &
    $1{,}888.0$ &
    $\mathbf{1{,}500.0 \pm 338.7}$ &
    $0.0 \pm 0.0$ &
    $-1{,}500.0$ \\

    Crazy Climber &
    $66{,}776.0$ &
    $\mathbf{67{,}505.0 \pm 31{,}738.8}$ &
    $0.0 \pm 0.0$ &
    $-67{,}505.0$ \\

    Demon Attack &
    $165.0$ &
    $\mathbf{217.0 \pm 99.6}$ &
    $142.0 \pm 95.9$ &
    $-75.0$ \\

    Freeway &
    $34.0$ &
    $\mathbf{0.2 \pm 0.4}$ &
    $0.0 \pm 0.0$ &
    $-0.2$ \\

    Frostbite &
    $1{,}316.0$ &
    $\mathbf{258.5 \pm 3.7}$ &
    $31.0 \pm 42.8$ &
    $-227.5$ \\

    Gopher &
    $8{,}240.0$ &
    $\mathbf{9{,}002.0 \pm 4{,}033.6}$ &
    $0.0 \pm 0.0$ &
    $-9{,}002.0$ \\

    Hero &
    $11{,}044.0$ &
    $\mathbf{7{,}747.5 \pm 94.1}$ &
    $0.0 \pm 0.0$ &
    $-7{,}747.5$ \\

    James Bond &
    $509.0$ &
    $\mathbf{445.0 \pm 113.4}$ &
    $70.0 \pm 49.7$ &
    $-375.0$ \\

    Kangaroo &
    $4{,}208.0$ &
    $\mathbf{3{,}610.0 \pm 491.9}$ &
    $0.0 \pm 0.0$ &
    $-3{,}610.0$ \\

    Krull &
    $8{,}413.0$ &
    $\mathbf{4{,}382.0 \pm 924.7}$ &
    $0.0 \pm 0.0$ &
    $-4{,}382.0$ \\

    Kung Fu Master &
    $26{,}182.0$ &
    $\mathbf{29{,}720.0 \pm 8{,}338.2}$ &
    $0.0 \pm 0.0$ &
    $-29{,}720.0$ \\

    Ms. Pac-Man &
    $2{,}673.0$ &
    $\mathbf{2{,}167.5 \pm 893.7}$ &
    $60.0 \pm 0.0$ &
    $-2{,}107.5$ \\

    Pong &
    $11.0$ &
    $\mathbf{18.7 \pm 1.6}$ &
    $-21.0 \pm 0.0$ &
    $-39.7$ \\

    Private Eye &
    $7{,}781.0$ &
    $\mathbf{100.0 \pm 0.0}$ &
    $-1{,}000.0 \pm 0.0$ &
    $-1{,}100.0$ \\

    Qbert &
    $4{,}522.0$ &
    $\mathbf{4{,}410.0 \pm 136.3}$ &
    $450.0 \pm 0.0$ &
    $-3{,}960.0$ \\

    Road Runner &
    $17{,}564.0$ &
    $\mathbf{10{,}320.0 \pm 2{,}731.0}$ &
    $0.0 \pm 0.0$ &
    $-10{,}320.0$ \\

    Seaquest &
    $525.0$ &
    $\mathbf{627.0 \pm 221.1}$ &
    $35.0 \pm 8.9$ &
    $-592.0$ \\

    Up N Down &
    $7{,}985.0$ &
    $\mathbf{5{,}012.0 \pm 625.6}$ &
    $1{,}121.0 \pm 529.1$ &
    $-3{,}891.0$ \\

    \bottomrule
  \end{tabular}
\end{table}

\FloatBarrier

\endgroup

\section{CGSReg Implementation Details}
\label{app:cgsreg-implementation-details}

In \mainCGSRegSectionRef{}, we defined \methodname{} for a single concept region.
Here we describe loss-scale matching, multiple masks, and the implementation
in each project.

\subsection{Loss Reduction and Multiple Masks}
\label{app:loss-reduction-and-multiple-masks}

For a concept mask $m$, we use the region-normalized reconstruction loss
\begin{equation}
\ell(m)
=
\frac{
\sum_p m_p (x_p-\hat{x}_p)^2
}{
\max(1,\sum_p m_p)
},
\end{equation}
where $p$ indexes image values after broadcasting the spatial mask across
channels. For multiple enabled masks, we first aggregate their independently
normalized losses:
\begin{equation}
\mathcal{L}_{\mathrm{region}}
=
\sum_{m \in \mathcal{M}_{\mathrm{enabled}}} \ell(m).
\end{equation}

We then match the reduction convention of the original image reconstruction
loss. If $\mathcal{L}_{\mathrm{img}}$ uses mean reduction over an image of
shape $H \times W \times C$, we set
$\mathcal{L}_{\mathrm{CGSReg}}=\mathcal{L}_{\mathrm{region}}$.
If $\mathcal{L}_{\mathrm{img}}$ uses sum reduction, we instead set
$\mathcal{L}_{\mathrm{CGSReg}}=HWC\,\mathcal{L}_{\mathrm{region}}$.
This keeps the scale of \methodname{} consistent with the image-loss
reduction used by each project.

Because each mask is normalized independently, its contribution does not
scale with its spatial size. The main experiments enable only the ball mask.

\subsection{Implementation in Each Project}
\label{app:cgsreg-implementation-by-project}

\autoref{tab:cgsreg-implementation-by-project} summarizes where \methodname{} is
applied in each project and how its scale is matched to the original image
loss.

\begin{table}[htbp]
  \centering
  \small
  \caption{\methodname{} implementation in each world-model project.}
  \label{tab:cgsreg-implementation-by-project}
  \begin{tabular}{@{}
    >{\raggedright\arraybackslash}p{0.18\linewidth}
    >{\raggedright\arraybackslash}p{0.36\linewidth}
    >{\raggedright\arraybackslash}p{0.36\linewidth}@{}}
    \toprule
    Project & Where \methodname{} is applied & Implementation note \\
    \midrule
    DreamerV3 &
    Decoded-image reconstruction from RSSM latents &
    Uses sum-reduction scaling to match the original image loss. \\

    DIAMOND &
    Denoising image loss &
    Uses the region-normalized masked loss directly. \\

    TWISTER &
    Decoded-image reconstruction from latent states &
    Uses the region-normalized masked loss directly. \\

    Simulus &
    Tokenizer image reconstruction &
    Applied only to tokenizer reconstruction, not token dynamics. \\

    STORM &
    Decoded-image reconstruction from predicted latents &
    Uses the region-normalized masked loss directly. \\
    \bottomrule
  \end{tabular}
\end{table}

\FloatBarrier
\section{SAM2 Segmentation Pipeline}
\label{app:sam2-segmentation-pipeline}

We augment the 100k-step Pong replay with object masks generated by
SAM2~\citep{ravi2024sam2}. Rather than manually prompting every replay
trajectory, we reuse a small set of manually prompted guiding frames. For each
replay chunk, the guiding frames are temporarily inserted, SAM2 propagates
their object masks, and the inserted frames are removed afterward. Long
trajectories are processed in chunks to limit GPU memory usage.

We generate three masks: \texttt{mask1} for the ball, \texttt{mask2} for the
opponent paddle, and \texttt{mask3} for the player paddle. The masks are
aligned with the original replay frames and stored as additional dataset
fields. The main experiments use only \texttt{mask1}.

\begin{algorithm}[h]
\caption{Guided SAM2 segmentation with reusable guiding frames}
\label{alg:guided-sam2-segmentation}
\begin{algorithmic}[1]
\small
\STATE \textbf{Input:} replay trajectories
$\{V_i\}_{i=1}^{N}$, guiding frames
$\{(g_j,P_j)\}_{j=1}^{K}$, and chunk length $L$
\STATE Initialize empty mask dataset $D_{\mathrm{mask}}$
\FOR{each replay trajectory $V_i=(x_1,\ldots,x_T)$}
    \FOR{each temporal chunk $C$ of $V_i$ with length $L$}
        \STATE Select guiding frames and prompts
        $\{(g_j,P_j)\}_{j\in\mathcal{J}}$
        \STATE Insert guiding frames into $C$ to obtain $C^+$
        \STATE Run SAM2 segmentation:
        $Y^+=\mathrm{SAM2}(C^+,\{P_j\}_{j\in\mathcal{J}})$
        \STATE Remove masks for the inserted guiding frames
        \STATE Align the remaining masks with the original frames in $C$
        \STATE Store the aligned masks in $D_{\mathrm{mask}}$
    \ENDFOR
\ENDFOR
\STATE Return the replay dataset augmented with concept masks
\end{algorithmic}
\end{algorithm}

\FloatBarrier
\section{Offline World-Model Training}
\label{app:offline-world-model-training}

The matched offline baseline and \methodname{} checkpoints are trained on the
same fixed 100k-step Pong replay collected from our DIAMOND reproduction. Only
world-model components are optimized; RL-component updates are disabled.

Because the projects use different training conventions, we match the baseline
and \methodname{} runs within each project by the number of world-model
gradient updates. \autoref{tab:offline-wm-training-budgets} summarizes the
budgets and optimized components.

\begin{table}[htbp]
  \centering
  \small
  \caption{Matched offline world-model training budgets. Update counts include
  only world-model optimization.}
  \label{tab:offline-wm-training-budgets}
  \begin{tabular}{@{}
    >{\raggedright\arraybackslash}p{0.18\linewidth}
    >{\raggedright\arraybackslash}p{0.30\linewidth}
    >{\raggedright\arraybackslash}p{0.38\linewidth}@{}}
    \toprule
    Project & World-model updates & Updated components \\
    \midrule
    DreamerV3 &
    25k &
    RSSM and image decoder \\

    DIAMOND &
    409.6k &
    Diffusion world model \\

    TWISTER &
    100k &
    Latent dynamics and image decoder \\

    Simulus &
    119k tokenizer + 115k dynamics &
    Image tokenizer and token dynamics \\

    STORM &
    100k &
    Latent dynamics and image decoder \\
    \bottomrule
  \end{tabular}
\end{table}

The update counts are converted from each project's native training schedule as
follows:

\begin{itemize}[leftmargin=*]
  \item \textbf{DreamerV3.} The Atari100K reproduction uses 100k environment
  steps, a train ratio of 256, batch size 16, and sequence length 64. Thus, the
  world model receives
  $100{,}000\times256/(16\times64)=25{,}000$ optimizer updates. The offline
  counter is set to 25.6M replay steps, which gives the same
  $25{,}600{,}000/(16\times64)=25{,}000$ updates.

  \item \textbf{DIAMOND.} Training runs for 1,000 epochs. The first epoch uses
  10,000 denoiser updates, and each of the remaining 999 epochs uses 400,
  yielding $10{,}000+999\times400=409{,}600$ updates.

  \item \textbf{TWISTER.} The native schedule contains 50 epochs with 2,000
  model steps per epoch. Each step invokes one world-model update, so both the
  reproduction and offline checkpoint use
  $50\times2{,}000=100{,}000$ updates.

  \item \textbf{Simulus.} Over 600 epochs, tokenizer training starts after
  epoch 5 and token-dynamics training starts after epoch 25; both use 200
  updates per active epoch. This gives $(600-5)\times200=119{,}000$ tokenizer
  updates and $(600-25)\times200=115{,}000$ dynamics updates. Actor--critic
  updates are disabled offline and are not counted.

  \item \textbf{STORM.} The native Atari100K schedule uses 100k sample steps
  with a train ratio of one, while the offline loop executes exactly 100k
  world-model-only iterations. This matches the number of optimizer updates,
  although the offline batch size is larger and therefore the total number of
  sampled sequences is not matched.
\end{itemize}

\FloatBarrier
\section{Unified Pixel-Space Zero-Shot MBRL Protocol}
\label{app:unified-pixel-space-zero-shot-mbrl-protocol}

For every frozen world model, we train a new policy using the same pixel-space
protocol summarized in \autoref{tab:unified-pixel-space-zero-shot-mbrl-in-pong-protocol}. World-model
parameters remain fixed, and no additional real-environment data are collected.

\begin{table}[htbp]
  \centering
  \small
  \caption{Unified pixel-space zero-shot MBRL protocol.}
  \label{tab:unified-pixel-space-zero-shot-mbrl-in-pong-protocol}
  \begin{tabular}{@{}
    >{\raggedright\arraybackslash}p{0.32\linewidth}
    >{\raggedright\arraybackslash}p{0.58\linewidth}@{}}
    \toprule
    Setting & Value \\
    \midrule
    Initial context &
    Sampled from the fixed replay dataset \\

    Policy input &
    $64\times64\times3$ RGB observations only \\

    Policy &
    Discrete-action on-policy actor--critic with an image encoder,
    categorical policy head, and value head \\

    Parallel WM environments &
    64 \\

    Rollout steps per update &
    15 \\

    Policy updates &
    20k \\

    Total WM environment steps &
    19.2M \\

    Maximum rollout horizon &
    512; reaching the horizon is treated as truncation \\

    Rewards &
    $-1,0,+1$ \\

    Continuous-reward handling &
    Threshold of $0.1$ before policy learning \\

    Real-ALE evaluation &
    20 reset seeds with deterministic action selection \\
    \bottomrule
  \end{tabular}
\end{table}

\FloatBarrier
\section{Ablations}
\label{app:ablations}

We study the sensitivity of \methodname{} to both the replay dataset and
regularization weight. We sweep
$\lambda_{\mathrm{CGSReg}}\in\{0,0.01,0.1,1.0\}$
using replay datasets collected from our DIAMOND and STORM reproductions.
All runs are evaluated with unified pixel-space zero-shot MBRL.

\begin{table}[htbp]
  \centering
  \small
  \caption{Replay-dataset and regularization-weight ablations.
  Each entry reports the mean $\pm$ standard deviation across
  three WM training seeds, with each seed evaluated over 20 real-ALE
  episodes. Bold indicates the best
  result within each row. Ties are not bolded.}
  \label{tab:replay-dataset-regularization-weight-ablations}
  \setlength{\tabcolsep}{3pt}
  \renewcommand{\arraystretch}{1.08}
  \begin{tabular}{@{}llrrrr@{}}
    \toprule
    Dataset &
    World model &
    $\lambda=0$ &
    $\lambda=0.01$ &
    $\lambda=0.1$ &
    $\lambda=1.0$ \\
    \midrule
    DIAMOND replay &
    DreamerV3 &
    $-19.2 \pm 3.1$ &
    $-21.0 \pm 0.0$ &
    $\mathbf{-13.4 \pm 1.7}$ &
    $-21.0 \pm 0.0$ \\

    &
    DIAMOND &
    $-9.6 \pm 4.5$ &
    $\mathbf{-1.0 \pm 4.2}$ &
    $-18.7 \pm 3.2$ &
    $-20.8 \pm 0.1$ \\

    &
    TWISTER &
    $-20.3 \pm 0.9$ &
    $-21.0 \pm 0.0$ &
    $-16.1 \pm 7.3$ &
    $\mathbf{-8.3 \pm 10.5}$ \\

    &
    Simulus &
    $-16.8 \pm 3.8$ &
    $\mathbf{-5.1 \pm 5.7}$ &
    $-8.0 \pm 6.0$ &
    $-20.8 \pm 0.4$ \\

    &
    STORM &
    $-21.0 \pm 0.0$ &
    $-21.0 \pm 0.0$ &
    $-21.0 \pm 0.0$ &
    $-21.0 \pm 0.0$ \\
    \midrule
    STORM replay &
    DreamerV3 &
    $-18.6 \pm 2.5$ &
    $\mathbf{-15.2 \pm 9.4}$ &
    $-20.2 \pm 0.7$ &
    $-21.0 \pm 0.0$ \\

    &
    DIAMOND &
    $-19.7 \pm 0.6$ &
    $-17.4 \pm 2.7$ &
    $\mathbf{-15.1 \pm 9.1}$ &
    $-20.3 \pm 1.1$ \\

    &
    TWISTER &
    $-21.0 \pm 0.0$ &
    $\mathbf{-17.2 \pm 3.4}$ &
    $-20.8 \pm 0.4$ &
    $-21.0 \pm 0.0$ \\

    &
    Simulus &
    $1.0 \pm 8.1$ &
    $\mathbf{2.8 \pm 0.9}$ &
    $-17.7 \pm 2.8$ &
    $-19.2 \pm 1.0$ \\

    &
    STORM &
    $-21.0 \pm 0.0$ &
    $-21.0 \pm 0.0$ &
    $-21.0 \pm 0.0$ &
    $-21.0 \pm 0.0$ \\
    \bottomrule
  \end{tabular}
\end{table}

For DreamerV3, DIAMOND, TWISTER, and Simulus, at least one nonzero weight
improves over $\lambda=0$ on both replay datasets. However, the best weight
varies across architectures and datasets, indicating that the effect of
\methodname{} is sensitive to its regularization strength. STORM remains
unchanged across all settings.

\FloatBarrier

\section{Dyna-Style Agents}
\label{app:dyna-style-training-and-agent-architectures}

The five evaluated agents share a broad Dyna-style training pattern:
real-environment data are collected into replay, the world model is updated
from replay, and the policy is improved using trajectories generated by the
world model. \autoref{alg:generic-dyna-style-mbrl} summarizes this common
structure.

\begin{algorithm}[htbp]
  \caption{Generic Dyna-style model-based reinforcement learning}
  \label{alg:generic-dyna-style-mbrl}
  \begin{algorithmic}[1]
    \STATE Initialize policy $\pi$, world model $M_\phi$, and replay buffer
    $\mathcal{D}$
    \FOR{training iteration $k = 1,2,\ldots$}
      \STATE Collect real-environment transitions with $\pi$ and add them to
      $\mathcal{D}$
      \STATE Update $M_\phi$ using data sampled from $\mathcal{D}$
      \STATE Generate trajectories with $M_\phi$ under $\pi$
      \STATE Improve $\pi$ using the generated trajectories
    \ENDFOR
    \STATE \textbf{return} $\pi$ and $M_\phi$
  \end{algorithmic}
\end{algorithm}

These agents differ not only in the space in which their world-model dynamics evolve, but also in the space in which the policy operates during their native MBRL. \autoref{tab:visual-wm-architecture-coverage} summarizes both.

\begin{table}[htbp]
  \centering
  \small
  \setlength{\tabcolsep}{3pt}
  \renewcommand{\arraystretch}{1.12}
  \caption{Architectural coverage of the five evaluated visual world-model
  agents. WM parameters exclude policy/value networks and fixed target/loss
  networks. Native policy input is the representation consumed by the policy
  during each project's native MBRL. For Simulus, the count includes the image
  tokenizer and token-dynamics model but excludes the frozen LPIPS network.}
  \label{tab:visual-wm-architecture-coverage}

  \begin{tabular}{@{}l
    >{\raggedright\arraybackslash}p{0.09\linewidth}
    >{\raggedright\arraybackslash}p{0.14\linewidth}
    >{\raggedright\arraybackslash}p{0.16\linewidth}
    >{\raggedright\arraybackslash}p{0.14\linewidth}
    >{\raggedright\arraybackslash}p{0.22\linewidth}@{}}
    \toprule
    Project &
    WM size &
    Prediction &
    Policy input &
    Backbone &
    Design \\
    \midrule

    DreamerV3 &
    $223.1$M &
    Latent state &
    RSSM latent state &
    RSSM &
    Recurrent latent dynamics with observation decoding \\

    DIAMOND &
    $10.3$M &
    Pixels &
    Generated pixels &
    Diffusion model &
    Action-conditioned diffusion in pixel space \\

    TWISTER &
    $41.3$M &
    Latent state &
    Latent state &
    Transformer &
    Transformer latent dynamics with AC-CPC \\

    Simulus &
    $16.8$M &
    Discrete visual tokens &
    Latent tokens &
    RetNet &
    Separate tokenizer and autoregressive token dynamics \\

    STORM &
    $11.8$M &
    Latent state &
    Latent state &
    Transformer &
    Transformer dynamics with stochastic latents \\

    \bottomrule
  \end{tabular}
\end{table}

\FloatBarrier
\subsection{DreamerV3}
\label{app:dreamer-introduction}

DreamerV3~\citep{hafner2023dreamerv3} is a representative latent
world-model-based reinforcement learning framework. It uses a recurrent
state-space model (RSSM) to model environment dynamics in a compact latent
space. The RSSM state contains a deterministic recurrent state and a
stochastic latent state represented by discrete categorical variables.
Conditioned on previous latent states and actions, the RSSM predicts future
latent states, while decoder networks reconstruct observations and predict
auxiliary signals such as rewards and continuation probabilities.

DreamerV3 performs reinforcement learning inside the learned world model
through latent imagination. Because dynamics prediction occurs in latent
space, policy optimization uses latent-state trajectories rather than pixel
observations. DreamerV3 stabilizes latent world-model training through design
choices including discrete categorical latent variables, KL balancing, and
symlog-based prediction targets.

These design choices enable DreamerV3 to perform strongly across diverse
reinforcement learning benchmarks. It is competitive on Atari100K, scales to
the full Atari benchmark, and applies to continuous-control tasks such as the
DeepMind Control Suite and long-horizon environments such as Minecraft.

\subsection{DIAMOND}
\label{app:diamond-introduction}

DIAMOND~\citep{alonso2024diamond} is a pixel-space visual model-based
reinforcement learning agent that uses a diffusion model for action-conditioned
future prediction. Given previous observations and actions, DIAMOND predicts
future frames by iteratively denoising noisy images. It therefore models future
pixels directly rather than learning an intermediate latent dynamics model.
This design enables high-fidelity visual rollouts, strong Atari100K results,
and an interactive neural game engine trained from Counter-Strike: Global
Offensive gameplay.

\subsection{TWISTER}
\label{app:twister-introduction}

TWISTER~\citep{burchi2025twister} adopts a latent-space world-model
architecture similar to STORM, using a Transformer-based dynamics model to
predict action-conditioned future latent states. It follows the Dreamer-style
latent-state modeling paradigm while replacing recurrent dynamics with
Transformer sequence modeling. Its main contribution is the
action-conditioned contrastive predictive coding (AC-CPC) objective, which
trains the model to predict and distinguish action-conditioned future latent
states and preserve information useful for long-horizon prediction.

\subsection{Simulus}
\label{app:simulus-introduction}

Simulus~\citep{cohen2025simulus} is the successor to
REM~\citep{cohen2024rem} and adopts a token-based visual world-modeling
paradigm. In REM, the visual world model is trained in two stages: a VQ-VAE
tokenizer~\citep{oord2017neural} first represents image observations as
discrete visual tokens, after which a RetNet-based autoregressive dynamics
model~\citep{sun2023retnet} learns next-token prediction conditioned on
actions. Simulus extends this framework with multi-modal tokenization and
several training improvements. Unlike the other evaluated agents, it models
future visual evolution through next-token prediction in a discrete token
space.

\subsection{STORM}
\label{app:storm-introduction}

STORM~\citep{zhang2023storm} follows the Dreamer-style latent world-model
design, where each latent state contains a deterministic representation and a
stochastic latent variable. Instead of the recurrent state-space model used by
DreamerV3, STORM employs a Transformer-based sequence model to predict
action-conditioned future latent trajectories. Its successor,
OC-STORM~\citep{ocstorm}, further incorporates object-centric representations
into the framework.

\FloatBarrier
\section{Diagnostic Study of World-Model Freezing in Dyna-Style Training}
\label{app:freeze-wm-diagnostic}

The strong RL performance of Dyna-style agents contrasts with the poor
standalone performance of their frozen world models. Although these agents can
achieve high returns during joint training, the corresponding frozen world
models often produce incorrect closed-loop rollouts and support weak
native zero-shot MBRL performance.

One possible explanation is that Dyna-style agents benefit from continual
world-model updating. During Dyna-style training, the policy and world model are
optimized together, while newly collected real-environment data continually
update the world model. The policy may therefore benefit from a model that is
useful near its current data distribution without requiring the model to become
a strong standalone simulator after freezing.

This hypothesis predicts that freezing the world model during Dyna-style
training should reduce final RL performance, especially when freezing occurs
earlier. To test this hypothesis, we freeze the world model at different
training stages and continue the remaining training process. For each project,
we follow its native Dyna-style training pipeline. After freezing,
real-environment interaction, data collection, policy learning, and evaluation
continue, while only world-model updates are disabled.

Let $T$ denote the nominal reproduction training budget. We compare four
settings: no freeze, freeze at $0.5T$, freeze at $0.75T$, and freeze at $1.0T$.
All runs continue until $1.5T$. This diagnostic follows each project's native
training and evaluation protocol and is therefore separate from the unified
pixel-space zero-shot MBRL protocol described in
\mainUnifiedZeroShotRef.

\begin{figure}[tbp]
  \centering
  \includegraphics[width=\linewidth]{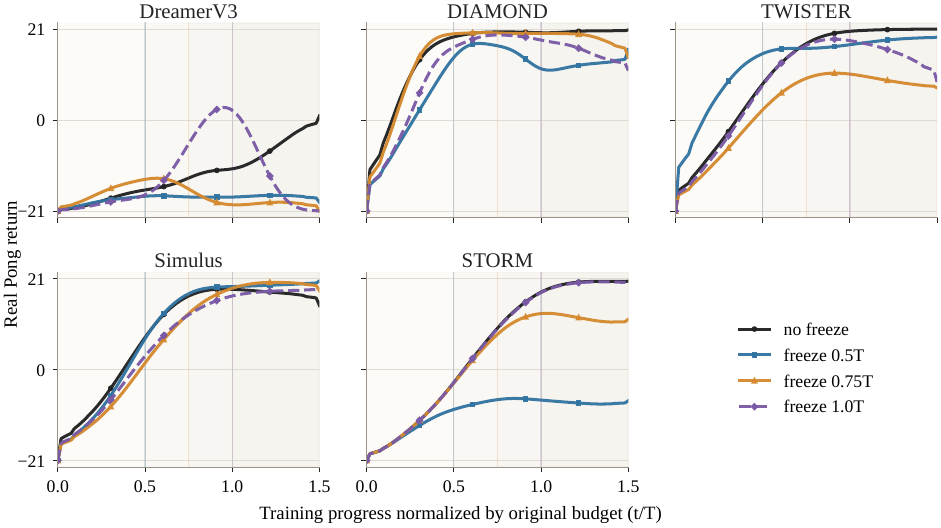}
  \caption{Scores after freezing world-model updates during Dyna-style
  training. The horizontal axis shows normalized progress $t/T$, and all runs
  continue to $1.5T$. Curves are smoothed and resampled for visual comparison;
  \autoref{tab:freeze-wm-final-returns} reports the final unsmoothed scores.}
  \label{fig:freeze-wm-diagnostic}
  \par\medskip

  \begin{minipage}{\linewidth}
    \centering
    \small
    \captionsetup{type=table}
    \caption{Final real-environment returns after freezing world-model
    updates during Dyna-style training. All runs continue to \(1.5T\), so the
    no-freeze scores are not identical to the reproduced-agent scores in
    \mainNativeZeroShotTableRef.}
    \label{tab:freeze-wm-final-returns}
    \begin{tabular}{@{}lrrrr@{}}
      \toprule
      World model & No freeze & Freeze 0.5T & Freeze 0.75T & Freeze 1.0T \\
      \midrule
      DreamerV3 & \(1.0\) & \(-19.0\) & \(-21.0\) & \(-21.0\) \\
      DIAMOND & \(20.8\) & \(16.2\) & \(14.5\) & \(11.4\) \\
      TWISTER & \(21.0\) & \(19.2\) & \(7.4\) & \(8.8\) \\
      Simulus & \(14.8\) & \(20.5\) & \(18.2\) & \(18.9\) \\
      STORM & \(20.4\) & \(-7.2\) & \(11.6\) & \(20.0\) \\
      \bottomrule
    \end{tabular}
  \end{minipage}
\end{figure}

\autoref{fig:freeze-wm-diagnostic} shows the performance trajectories after
freezing world-model updates. \autoref{tab:freeze-wm-final-returns} reports the final
real-environment returns.

The results provide partial support for the hypothesis. Freezing the world
model degrades performance in DreamerV3 and DIAMOND, with a weaker effect in
TWISTER, but not in Simulus; STORM shows non-monotonic behavior. Moreover, the
effect is not consistently stronger for earlier freezing. Continual world-model
updating can therefore contribute to Dyna-style performance, but it does not
fully explain the agent--world-model gap.

Other factors, such as robustness to policy shifts, long-horizon consistency,
and action-conditioned prediction, may also contribute and remain important
directions for future work.

\FloatBarrier

\end{document}